\documentclass[11pt,a4paper]{article}

\usepackage{graphicx} %$B2hA|A^F~MQ(B
\usepackage{bm, amsmath, amssymb, amsthm, natbib, accents}
\usepackage{algorithm}
\usepackage{algpseudocode, setspace}% http://ctan.org/pkg/algorithmicx

\begin{document}

\title{Empirical Bayes Matrix Completion}
\author{Takeru MATSUDA$^{\displaystyle 1}$ and Fumiyasu KOMAKI$^{\displaystyle 1,2}$\\ \\
        $^{\displaystyle 1}$Department of Mathematical Informatics\\
        Graduate School of Information Science and Technology\\
        The University of Tokyo\\
        $^{\displaystyle 2}$RIKEN Brain Science Institute\\
        \texttt{\normalsize \{matsuda,komaki\}@mist.i.u-tokyo.ac.jp}}

\maketitle

\begin{abstract}%
We develop an empirical Bayes (EB) algorithm for the matrix completion problems.
The EB algorithm is motivated from the singular value shrinkage estimator for matrix means by \cite{Efron72}.
Since the EB algorithm is derived as the EM algorithm applied to a simple model, it does not require heuristic parameter tuning other than tolerance.
Numerical results demonstrated that the EB algorithm achieves a good trade-off between accuracy and efficiency compared to existing algorithms 
and that it works particularly well when the difference between the number of rows and columns is large.
Application to real data also shows the practical utility of the EB algorithm.
\end{abstract}

\section{Introduction}
In various applications, we encounter problems of estimating the unobserved entries of a matrix from the observed entries.
For example, in the famous Netflix problem, we have a matrix of movie ratings by users 
and aim to predict the preference for movies of each user for recommendation.
This problem is called the matrix completion problem and many studies have investigated its theoretical properties \citep{Candes,Recht} 
and developed efficient algorithms \citep{Srebro,Cai,Keshavan,Mazumder}.
%and developed efficient algorithms such as the SVT algorithm \citep{Cai}, the SOFT-IMPUTE algorithm \citep{Mazumder}, 
%the OPTSPACE algorithm \citep{Keshavan}, and the MMMF algorithm \citep{Srebro}.

In the matrix completion problems, the low-rank property of the underlying matrix plays a central role.
For example, in the Netflix problem, the rank is interpreted as the number of latent factors in the movie preference and it is believed to be small.
Indeed, existing matrix completion algorithms succeed in estimating the unobserved entries by assuming the low-rankness.
Note that low rank matrices have sparse singular values since the rank of a matrix is equal to the number of its nonzero singular values.
The sum of singular values of a matrix is called the trace norm or nuclear norm, which is employed by many existing algorithms for regularization.

In practice, the data matrix often contains observation noise and we aim to recover the true underlying matrix.
If the data matrix is fully observed with the Gaussian observation noise, 
then the matrix completion problem reduces to the estimation of the mean matrix parameter of a matrix-variate normal distribution.
For this problem, \cite{Efron72} developed an empirical Bayes estimator 
and proved that it is minimax and dominates the maximum likelihood estimator under the Frobenius loss.
Later, \cite{Stein74} pointed out that this estimator shrinks the singular values of the observed matrix for each.
Therefore, this estimator performs well when the true value of the mean matrix parameter has low rank.
Based on this idea, \cite{Matsuda} developed singular value shrinkage priors as a natural generalization of the Stein prior.
The singular value shrinkage priors are superharmonic and the Bayes estimators based on them are minimax estimators with similar properties to the Efron--Morris estimator.

In this study, we develop an empirical Bayes (EB) algorithm for matrix completion.
The EB algorithm is a natural extension of the Efron--Morris estimator.
Since the EB algorithm is essentially the EM algorithm applied to a simple model, it does not require heuristic parameter tuning other than tolerance.
%Unlike existing methods, the proposed method does not require heuristic parameter tuning.
Numerical experiments demonstrate the effectiveness of the EB algorithm compared with existing algorithms.
Specifically, the EB algorithm works well when the difference between the number of rows and columns is large. 
Application to real data also shows the practical utility of the EB algorithm.

This paper is organized as follows.
Section 2 reviews the previous results on the empirical Bayes estimation of matrix means.
Section 3 provides details of the EB algorithm.
Section 4 presents the results of the numerical experiments while Section 5 applies the EB algorithm to real data.

\section{Empirical Bayes Estimation of Matrix Means}
In this section, we review the empirical Bayes estimator by \cite{Efron72} for the mean matrix parameter of a matrix-variate normal distribution.
We extend this estimator to matrix completion problems in the next section.

Suppose that we have a matrix observation $Y=(y_1, \cdots, y_p)^{\top} \in \mathbb{R}^{p \times q}$, 
whose row vectors $y_i^{\top} \in \mathbb{R}^q$ have the distribution $y_i \sim {\rm N}_q (m_i, I_q)$ independently, 
where $I_q$ is the $q$-dimensional identity matrix.
Here, $m_i^{\top} \in \mathbb{R}^q$ is the $i$-th row vector of $M=(m_1, \cdots, m_p)^{\top} \in \mathbb{R}^{p \times q}$.
In the notation of \cite{Dawid}, this situation is denoted as $Y \sim {\rm N}_{p,q} (M, I_p, I_q)$.
We assume $p \geq q$.
We consider estimation of $M$ under the Frobenius loss:
\begin{equation*}
	L(M,\hat{M}(Y)) = \| \hat{M}(Y) - M \|_{{\rm F}}^2 = \sum_{i=1}^p \sum_{j=1}^q (\hat{M}_{i j} (Y) - M_{i j})^2.
\end{equation*}

Let $Y = U \Lambda V^{\top}$ be the singular value decomposition of a matrix $Y$, 
where $U \in O(p)$, $V \in O(q)$, and $\Lambda = \left[ {\rm diag} (\sigma_1(Y), \ldots, \sigma_q(Y)) \; O_{q,p-q} \right]^{\top}$.
Here, $O(k)$ is the $k$-dimensional orthogonal group, $O_{lm}$ is the zero matrix of size $l \times m$, and
$\sigma_1(Y) \geq \cdots \geq \sigma_q(Y) \geq 0$ are the singular values of $Y$.
Similarly, let $\hat{M} = \hat{U} \hat{\Lambda} \hat{V}^{\top}$ be the singular value decomposition of an estimator $\hat{M}$ of $M$, 
where $\hat{U} \in O(p)$, $\hat{V} \in O(q)$, $\hat{\Lambda} = \left[ {\rm diag} (\sigma_1(\hat{M}), \cdots, \sigma_q(\hat{M})) \; O_{q,p-q} \right ]^{\top}$, and
$\sigma_1(\hat{M}) \geq \cdots \geq \sigma_q(\hat{M}) \geq 0$ are the singular values of $\hat{M}$.

\cite{Efron72} proposed the estimator
\begin{equation}
	\hat{M}_{{\rm EM}} = Y \left( I_q-(p-q-1) (Y^{\top} Y)^{-1} \right) \label{EM_estimator}
\end{equation}
and proved that it is minimax and dominates the maximum likelihood estimator $\hat{M}=Y$ when $p-q-1>0$.
When $q=1$, the Efron--Morris estimator $\hat{M}_{{\rm EM}}$ coincides with the James--Stein estimator.
\cite{Stein74} pointed out that $\hat{M}_{{\rm EM}}$ can be represented in the singular value decomposition form as follows:
\begin{equation*}
	\sigma_i(\hat{M}_{{\rm EM}}) = \left( 1 - \frac{p-q-1}{\sigma_i(Y)^2} \right) \sigma_i(Y) \quad (i=1, \ldots, q),
\end{equation*}
\begin{equation*}
	\hat{U} = U, \quad \hat{V} = V.
\end{equation*}
Therefore, $\hat{M}_{{\rm EM}}$ shrinks the singular values of $Y$ for each and preserves the singular vectors of $Y$.

The Efron--Morris estimator $\hat{M}_{{\rm EM}}$ in \eqref{EM_estimator} was derived as an empirical Bayes estimator based on the following hierarchical model:
\begin{equation*}
	M \sim {\rm N}_{p,q} (0, I_p, \Sigma),
\end{equation*}
\begin{equation*}
	Y \mid M \sim {\rm N}_{p,q} (M, I_p, I_q).
\end{equation*}
The above model assumes that each row vector $m_i^{\top} \in \mathbb{R}^q$ of $M=(m_1, \cdots, m_p)^{\top} \in \mathbb{R}^{p \times q}$ 
has the distribution $m_i \sim {\rm N}_q (0, \Sigma)$ independently.
If $\Sigma$ is given, then the Bayes estimator of $M$ is written as
\begin{equation}
	\hat{M}^{\Sigma} (Y)=Y (I_q-(I_q+\Sigma)^{-1}). \label{Best}
\end{equation}
To obtain an empirical Bayes estimator, we estimate $\Sigma$ from $Y$.
Since the marginal distribution of $Y$ is $Y \sim {\rm N}_{p,q} (0, I_p, I_q+\Sigma)$, 
the marginal distribution of $Y^{\top} Y$ is $Y^{\top} Y \sim W_q (p, I_q+\Sigma)$.
From the property of the Wishart distribution, we have ${\rm E} (Y^{\top} Y)^{-1} = (p-q-1)^{-1} (I_q+\Sigma)^{-1}$.
Thus, we can estimate $\Sigma$ by $(I_q+\hat{\Sigma})^{-1}=(p-q-1) (Y^{\top} Y)^{-1}$.
By substituting this estimate into \eqref{Best}, the Efron--Morris estimator $\hat{M}_{{\rm EM}}$ in \eqref{EM_estimator} is obtained.
We note that $\hat{M}_{{\rm EM}}$ is not a generalized Bayes estimator.

Recently, \cite{Matsuda} developed the singular value shrinkage prior 
\begin{equation*}
	\pi_{{\rm SVS}} (M) = \det (M^{\top} M)^{-(p-q-1)/2}
\end{equation*}
and proved its superharmonicity.
When $q=1$, the singular value shrinkage prior $\pi_{{\rm SVS}}$ coincides with the Stein prior $\pi_{{\rm S}}(\mu) = \| \mu \|^{2-p}$ \citep{Stein74}.
The generalized Bayes estimators based on the singular value shrinkage priors are minimax and have similar properties to the Efron--Morris estimator $\hat{M}_{{\rm EM}}$ in \eqref{EM_estimator}.
This is an extension of the relationship between the James--Stein estimator and the Stein prior.
We note that the singular value shrinkage prior has the integral representation
\begin{equation*}
%	\pi_{{\rm SVS}} (M) = \frac{2^{m(m+1)/2} \pi^{mn/2}}{\Gamma_m \left( \frac{n-m-1}{2} \right)} \int p(M \mid I_n, \Sigma) {\rm d} \Sigma,
	\pi_{{\rm SVS}} (M) \propto \int p(M \mid I_p, \Sigma) {\rm d} \Sigma,
\end{equation*}
where $p(M \mid I_p, \Sigma)$ is the probability density function of $M \sim {\rm N}_{p,q} (0, I_p, \Sigma)$ and 
${\rm d} \Sigma$ is the Lebesgue measure on the space of $q \times q$ positive-semidefinite matrices.
This can be confirmed by the calculation of the normalization constant in the inverse-Wishart distribution.

\section{The EB Algorithm}
In this section, we propose the empirical Bayes (EB) algorithm for the matrix completion problems.
This algorithm is motivated from the Efron--Morris estimator $\hat{M}_{{\rm EM}}$ in \eqref{EM_estimator}.

%We formulate the matrix completion problem as follows.
We assume that the data matrix $Y \in \mathbb{R}^{p \times q}$ has the distribution $Y \sim {\rm N}_{p,q} (M, \sigma^2 I_p, I_q)$, 
where $p \geq q$ and $\sigma^2$ is an unknown variance.
Namely, each row vector $y_i^{\top} \in \mathbb{R}^q$ of $Y=(y_1, \cdots, y_p)^{\top} \in \mathbb{R}^{p \times q}$ 
has the distribution $y_i \sim {\rm N}_q (m_i, I_q)$ independently, 
where $m_i^{\top} \in \mathbb{R}^q$ is the $i$-th row vector of $M=(m_1, \cdots, m_p)^{\top} \in \mathbb{R}^{p \times q}$.
We observe only part of the entries of $Y$.
Let $\Omega \subset \{1, \cdots, p \} \times \{ 1, \cdots, q \}$ be the set of indices of the observed entries 
and $\Omega_i = \{ j \mid (i,j) \in \Omega \} \ (i=1,\cdots,p)$ be the set of indices of the observed entries in the $i$-th row.
We denote the observed entries of $Y$ by $Y_{\Omega}$.
We denote the submatrix of a matrix $A$ with row indices $R$ and column indices $C$ as $A[R;C]$.
For example, if
\begin{equation*}
	A = \begin{pmatrix} 1 & 2 & 3 \\ 4 & 5 & 6 \\ 7 & 8 & 9 \end{pmatrix}, \quad R = \{ 1, 2 \}, \quad C = \{ 1, 3 \},
\end{equation*}
then we have
\begin{equation*}
	A[R; C] = \begin{pmatrix} 1 & 3 \\ 4 & 6 \end{pmatrix}.
\end{equation*}

Our goal is to estimate $M$ from the observed entries $Y_{\Omega}$ of $Y$.
We tackle this problem using an empirical Bayes approach based on the following hierarchical model:
\begin{equation}
	M \sim {\rm N}_{p,q} (0, I_p, \Sigma), \label{gen_model}
\end{equation}
\begin{equation}
	Y \mid M \sim {\rm N}_{p,q} (M, \sigma^2 I_p, I_q). \label{obs_model}
\end{equation}
Namely, we assume that each row vector $m_i^{\top} \in \mathbb{R}^q$ of $M=(m_1, \cdots, m_p)^{\top} \in \mathbb{R}^{p \times q}$ 
has the distribution $m_i \sim {\rm N}_q (0, \Sigma)$ independently.
Note that the Efron--Morris estimator $\hat{M}_{{\rm EM}}$ in \eqref{EM_estimator} was also derived from this model with $\sigma^2=1$.
%where the estimate $(I_q+\hat{\Sigma})^{-1}=(p-q-1) (Y^{\top} Y)^{-1}$ is plugged into the Bayes estimator $\hat{M}(Y)=Y (I_p-(I_p+\Sigma)^{-1})$.
Here, since only part of the entries of $Y$ are observed, 
we use the EM algorithm \citep{EMalgo} to estimate the hyperparameters $\Sigma$ and $\sigma^2$.
As a result, the EB algorithm is described as Algorithm \ref{EBalgo}.
Derivation of the EB algorithm is given in Appendix B.
Empirically, this algorithm converges in less than 20 iterations for most cases. % when $\varepsion = 10^{-3}$.
We note that the log-likelihood $\log p(Y_{\Omega} \mid \Sigma, \sigma^2)$ is obtained as
\begin{align*}
	\log p(Y_{\Omega} \mid \Sigma, \sigma^2) = & -\frac{|\Omega|}{2} \log(2 \pi) -\frac{1}{2} \sum_{i=1}^p \log \det (\sigma^2 I_{|\Omega_i|+\Sigma[\Omega_i,\Omega_i]}) \\
	& \quad - \frac{1}{2} \sum_{i=1}^p Y [\{ i \}; \Omega_i] (\sigma^2 I_{|\Omega_i|+\Sigma[\Omega_i,\Omega_i]})^{-1} Y [\{ i \}; \Omega_i]^{\top}.
\end{align*}
Similarly to the SOFT-IMPUTE algorithm by \cite{Mazumder},% also iteratively imputes the missing entries of $Y$.
the EB algorithm can be viewed as iteratively imputing the missing entries of $Y$, 
although we are updating not $Y$ but $M$ strictly speaking.

\begin{algorithm}
\setstretch{1.4}
\caption{EB algorithm}\label{EBalgo}

\textbf{Input:} set of observation indices $\Omega$, observed entries $Y_{\Omega}$, initial value $\sigma_0^2$, and tolerance $\varepsilon_1, \varepsilon_2$

\textbf{Output:} $\hat{M}$

\textbf{Description:} Estimate $M$ from $Y_{\Omega}$ based on the model \eqref{gen_model} and \eqref{obs_model}

\vspace{0.15in}

\begin{algorithmic}[1]
	\State{Initialize $M_{{\rm old}} \in \mathbb{R}^{p \times q}$ by $(M_{{\rm old}})_{i j} = \begin{cases} Y_{i j} & ((i,j) \in \Omega) \\ 0 & ((i,j) \not\in \Omega) \end{cases}$}
	\State{Initialize $\Sigma_{{\rm old}} = ({M_{{\rm old}}}^{\top} M_{{\rm old}})/p$}
	\State{Initialize $(\sigma^2)_{{\rm old}} = \sigma_0^2$}
	\While{true}
		\For{$i = 1 \, \cdots \, p$}
			\State{Set $P_i = O_{q,q}$ ($q \times q$ zero matrix)}
			\State{Set $P_i [\Omega_i,\Omega_i] = (\sigma^2_{{\rm old}} I_{| \Omega_i |} + \Sigma_{{\rm old}} [\Omega_i,\Omega_i])^{-1}$}
			\State{Set $R_i=\Sigma_{{\rm old}} - \Sigma_{{\rm old}} {P}_i \Sigma_{{\rm old}}$}
			\State{Set $b_i \in \mathbb{R}^q$ by $(b_i)_j = \begin{cases} Y_{ij} & (j \in \Omega_i) \\ 0 & (j \not\in \Omega_i) \end{cases}$}
%			\State{Set $m_i=R_i b_i$}
		\EndFor
		\State{Set $M_{{\rm new}} = (\sigma^2_{{\rm old}})^{-1} (R_1 b_1,\cdots,R_p b_p)^{\top}$}
		\State{Set $\Sigma_{{\rm new}} = \left( M_{{\rm new}} M_{{\rm new}}^{\top} + \sum_{i=1}^p R_i \right)/p$}
		\State{Set $\sigma^2_{{\rm new}} = \sum_{(i,j) \in \Omega} \left( (Y_{ij}-(M_{{\rm new}})_{ij})^2 + (R_i)_{jj} \right) / | \Omega |$}
		\If{$\log p(Y_{\Omega} \mid \Sigma_{{\rm new}}, \sigma^2_{{\rm new}}) - \log p(Y_{\Omega} \mid \Sigma_{{\rm old}}, \sigma^2_{{\rm old}}) < \varepsilon_1$}
			\State{Set $\hat{M} = M_{{\rm new}}$ and exit}
		\EndIf
		\If{$\| M_{{\rm new}}-M_{{\rm old}} \|_{{\rm F}}^2 / \| M_{{\rm old}} \|_{{\rm F}}^2 < \varepsilon_2$}
			\State{Set $\hat{M} = M_{{\rm new}}$ and exit}
		\EndIf
		\State{Set $M_{{\rm old}} = M_{{\rm new}}$}
		\State{Set ${\Sigma}_{{\rm old}} = {\Sigma}_{{\rm new}}$}
		\State{Set $\sigma^2_{{\rm old}} = \sigma^2_{{\rm new}}$}
	\EndWhile
\end{algorithmic}
\end{algorithm}

%The SOFT-IMPUTE algorithm by \cite{Mazumder} also iteratively imputes the missing entries of $Y$.
%Whereas the SOFT-IMPUTE algorithm employs cross-validation for the selection of a regularization parameter,
%our EB algorithm determines the strength of regularization by maximizing the marginal likelihood $p(Y_{\Omega} \mid \Sigma, \sigma^2)$.

%In practice, the variance of observation noise $\sigma^2$ is unknown. % although the Efron--Morris estimator assumes it is known.
%The EB algorithm determines $\sigma^2$ in a data-driven manner, since it is based on the EM algorithm.
%We note that the SVT algorithms require the manual setting of the magnitude of observation noise:
%e.g., the parameter $E_{ij}$ in SVT. % or the parameter $\delta$ in SOFT-IMPUTE.

\section{Numerical Experiments}
In this section, we investigate the performance of the EB algorithm by numerical experiments.
The EB algorithm is compared with the SVT algorithm by \cite{Cai}, the SOFT-IMPUTE algorithm by \cite{Mazumder}, and the OPTSPACE algorithm by \cite{Keshavan}.
For these existing algorithms, we use the MATLAB codes provided by the authors online. 

The SVT algorithm \citep{Cai} solves the following optimization problem:
\begin{align*}
	\underset{\hat{M}}{{\rm minimize}} & \quad \| \hat{M} \|_{*} \\
	{\rm subject\ to} & \quad |Y_{ij}-\hat{M}_{ij}| \leq E_{ij}, \quad (i,j) \in \Omega,
\end{align*}
where $\| \cdot \|_{*}$ denotes the nuclear norm and $E_{ij}$ is a tolerance parameter.
In the same manner as the original paper, we set $E_{i j}$ equal to the standard deviation of the observation noise $\sigma$. % 5.2.2 in Candes
We adopt the default settings of the algorithm parameters: $\tau = 5 \sqrt{pq}$, $\delta = 1.2pq/|\Omega|$, $k_{\max} = 1000$, $\epsilon = 10^{-4}$.

The SOFT-IMPUTE algorithm \citep{Mazumder} solves the following optimization problem:
\begin{align*}
	\underset{\hat{M}}{{\rm minimize}} & \quad \| \hat{M} \|_{*} \\
	{\rm subject\ to} & \quad \sum_{(i,j) \in \Omega} (Y_{ij}-\hat{M}_{ij})^2 \leq \delta,
\end{align*}
which is rewritten as
\begin{align*}
	\underset{\hat{M}}{{\rm minimize}} \quad \frac{1}{2} \sum_{(i,j) \in \Omega} (Y_{ij}-\hat{M}_{ij})^2 + \lambda \| \hat{M} \|_{*},
\end{align*}
where $\lambda \geq 0$ is a regularization parameter.
In the same manner as the original paper, we select $\lambda$ from $K$ candidate values 
by cross-validation using $\eta$ \% of the observed entries as the validation set.
We adopt the default settings of the algorithm parameters: $K=20$, $\eta=20$, $\epsilon = 10^{-4}$.

The OPTSPACE algorithm \citep{Keshavan} achieves matrix completion via spectral techniques and manifold optimization. 
Here, we use the option of guessing the rank $r$ from data.
We adopt the default setting of algorithm parameters: maximum number of iterations = $50$, tolerance = $10^{-6}$.

We consider the same experimental setting with \cite{Mazumder}.
%We sample $U \sim {\rm N}_{p,r} (0,I_p,I_r), V \sim {\rm N}_{r,q} (0,I_r,I_q)$ independently and set $M = U V$.
We generate $U \in \mathbb{R}^{p \times r}$ and $V \in \mathbb{R}^{r \times q}$ 
whose entries are sampled from the standard normal distribution ${\rm N} (0,1)$ independently and put $M = U V$.
Here, $r$ denotes the rank of $M$.
Then, we generate $Y=M+E$, where the entries of $E$ are sampled from ${\rm N} (0,\sigma^2)$ independently.
%This process corresponds to the observation model \eqref{obs_model}.
The indices of the observed entries $\Omega$ are randomly sampled over $pq$ indices of the matrix.
We evaluate the accuracy of the matrix completion algorithms by the normalized error for the overall matrix
\begin{equation*}
	{\rm error1} := \frac{\| \hat{M}-M \|_{{\rm F}}}{\| M \|_{{\rm F}}},
\end{equation*}
and the normalized error for the unobserved entries
\begin{equation*}
	{\rm error2} := \frac{(\sum_{(i,j) \not\in \Omega} (\hat{M}_{ij}-M_{ij})^2)^{1/2}}{(\sum_{(i,j) \not\in \Omega} M_{ij}^2)^{1/2}}.
\end{equation*}
%In practice, ${\rm error}_2$ measures the generalization error is more important than ${\rm error}_1$.
We also compare the efficiency of the matrix completion algorithms by the computation time in seconds.

\begin{figure}
 \begin{minipage}{0.5\hsize}
 (a)
 \begin{center}
	\includegraphics[width=7.5cm]{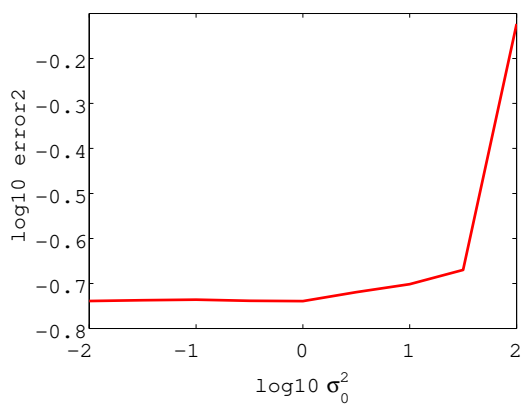}
 \end{center}
 \end{minipage}
 \begin{minipage}{0.5\hsize}
 (b)
 \begin{center}
	\includegraphics[width=7.5cm]{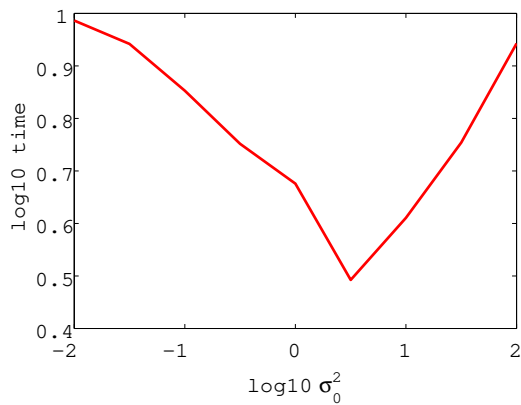}
 \end{center}
 \end{minipage}
	\caption{Plot of (a) error2 and (b) computation time (in seconds) as a function of $\sigma_0^2$, where $p=1000$, $q=100$, $r=10$, $| \Omega |/(pq)=0.5$ and $\sigma^2=1$.}
	\label{init_sigma2_plot}
\end{figure}

For the EB algorithm, we set $\varepsilon_1=10^{-3}$ and $\varepsilon_2=10^{-4}$.
Also, we set $\sigma_0^2=\sigma^2$ except for Figure \ref{init_sigma2_plot}.
In Figure \ref{init_sigma2_plot}, we investigate how the selection of $\sigma_0^2$ affects the performance of the EB algorithm.

Table \ref{tab_1000_100_10_50_1} shows the results when $p=1000$, $q=100$, $r=10$, $\sigma^2=1$, and $| \Omega |/(pq) = 0.5$.
We present error1, error2, and the computation time averaged over 100 simulations.
EB and OPTSPACE have the least error$_2$, whereas EB is faster than OPTSPACE.
SOFT-IMPUTE takes the least computation time, whereas its error$_2$ is almost twice as large as those of EB and OPTSPACE.
Therefore, EB achieves a good trade-off between accuracy and efficiency under this setting.

\begin{table}[htbp]
	\centering
	\begin{tabular}{|c|c|c|c|}
	\hline
	 & error1 & error2 & time \\ \hline
	EB & 0.21 & 0.18 & 4.63 \\ \hline
	SVT & 0.28 & 0.31 & 5.12 \\ \hline
	SOFT-IMPUTE & 0.28 & 0.31 & 2.75 \\ \hline
	OPTSPACE & 0.16 & 0.17 & 8.06 \\ \hline
	\end{tabular}
	\caption{Performance of the matrix completion algorithms ($p=1000$, $q=100$, $r=10$, $| \Omega |/(pq) = 0.5$, and $\sigma^2=1$)}
	\label{tab_1000_100_10_50_1}
\end{table}

Now, we investigate the dependence of the performance of each algorithm on the size of the matrix $(p, q)$, 
the true rank $r$, the proportion of observed entries $| \Omega |/(pq)$, and the observation noise variance $\sigma^2$.
We present error2 and the computation time averaged over 100 simulations for each setting.
The qualitative behavior of error1 was almost the same as that of error2.

\begin{figure}
 \begin{minipage}{0.5\hsize}
 (a)
 \begin{center}
	\includegraphics[width=7.5cm]{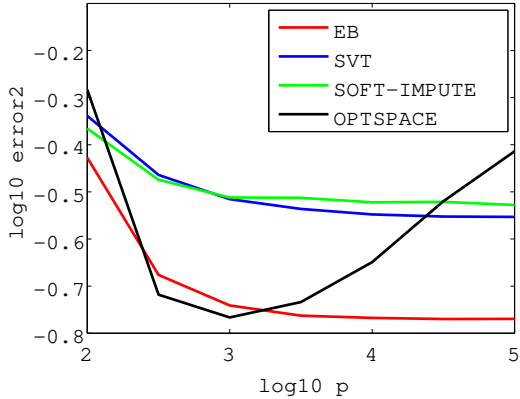}
 \end{center}
 \end{minipage}
 \begin{minipage}{0.5\hsize}
 (b)
 \begin{center}
	\includegraphics[width=7.5cm]{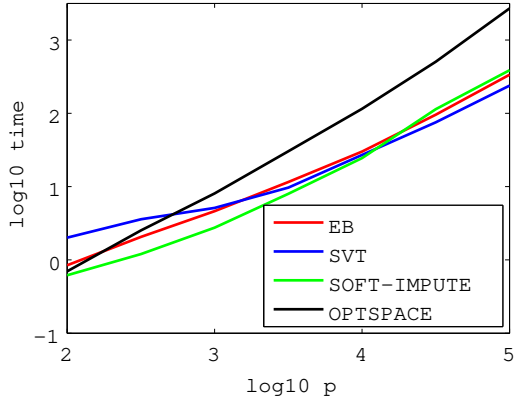}
 \end{center}
 \end{minipage}
%	\caption{Plot of (a) error2 and (b) computation time (in seconds) with respect to $p$, where $q=100$, $r=10$, $\sigma^2=1$, and $| \Omega |/(pq)=0.5$. red: the proposed method, blue: SVT, green: SOFT-IMPUTE.}
	\caption{Plot of (a) error2 and (b) computation time (in seconds) as a function of $p$, where $q=100$, $r=10$, $\sigma^2=1$, and $| \Omega |/(pq)=0.5$.}% solid line: EB, dashed line: SVT, dotted line: SOFT-IMPUTE.}
	\label{long_plot}
\end{figure}

Figure \ref{long_plot} plots error2 and the computation time as a function of $p \in [10^2, 10^5]$, 
where $q=100$, $r=10$, $\sigma^2=1$, and $| \Omega |/(pq)=0.5$.
Here, error2 decreases with $p$ for EB, SVT, and SOFT-IMPUTE and the computation time increases with $p$ for all algorithms.
EB has the best accuracy, whereas its computation time is almost the same as those of SVT and SOFT-IMPUTE.
The accuracy of OPTSPACE is almost the same as EB when $p \leq 10^3$ but becomes worse when $p > 10^3$.
Also, the computation time of OPTSPACE grows with $p$ faster than the other algorithms.
%The computation time is almost the same for three algorithms and it is approximately $O(p^{5/6})$.

\begin{figure}
 \begin{minipage}{0.5\hsize}
 (a)
 \begin{center}
	\includegraphics[width=7.5cm]{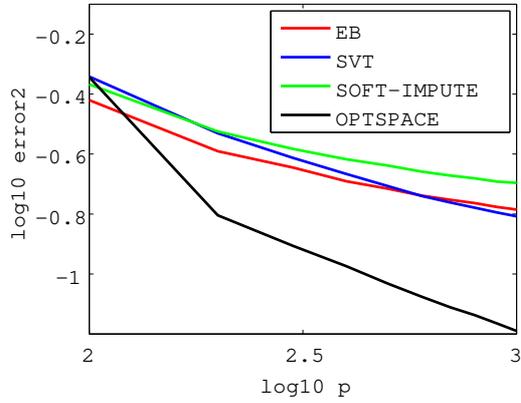}
 \end{center}
 \end{minipage}
 \begin{minipage}{0.5\hsize}
 (b)
 \begin{center}
	\includegraphics[width=7.5cm]{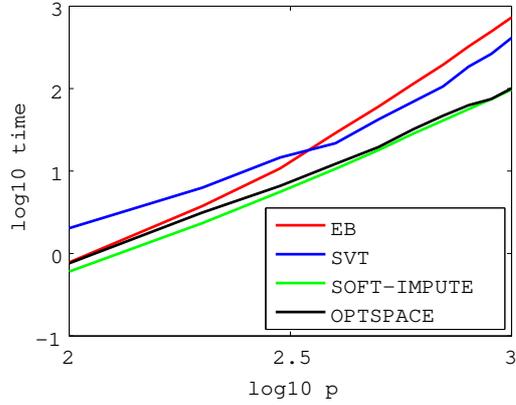}
 \end{center}
 \end{minipage}
%	\caption{Plot of (a) error2 and (b) computation time (in seconds) with respect to $p$, where $q=p$, $r=10$, $\sigma^2=1$, and $| \Omega |/(pq)=0.5$. red: the proposed method, blue: SVT, green: SOFT-IMPUTE.}
	\caption{Plot of (a) error2 and (b) computation time (in seconds) as a function of $p$, where $q=p$, $r=10$, $\sigma^2=1$, and $| \Omega |/(pq)=0.5$.}
	\label{square_plot}
\end{figure}

Figure \ref{square_plot} plots error2 and the computation time as a function of $p \in [10^2, 10^3]$, 
where $q=p$, $r=10$, $\sigma^2=1$, and $| \Omega |/(pq)=0.5$.
Here, we are considering the case of a square matrix.
For all algorithms, error2 decreases with $p$ and the computation time increases with $p$.
OPTSPACE has the best accuracy and efficiency.
EB has almost the same accuracy with SVT and SOFT-IMPUTE, whereas its computation time grows with $p$ a little faster than the others.

\begin{figure}
 \begin{minipage}{0.5\hsize}
 (a)
 \begin{center}
	\includegraphics[width=7.5cm]{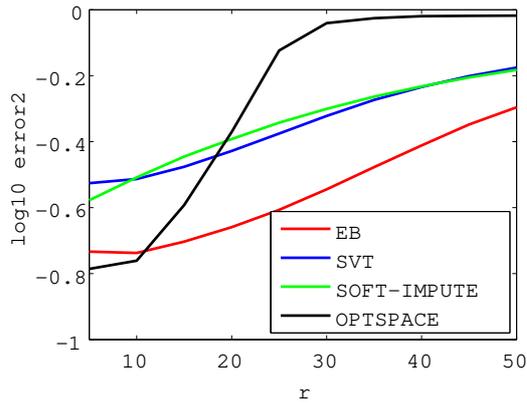}
 \end{center}
 \end{minipage}
 \begin{minipage}{0.5\hsize}
 (b)
 \begin{center}
	\includegraphics[width=7.5cm]{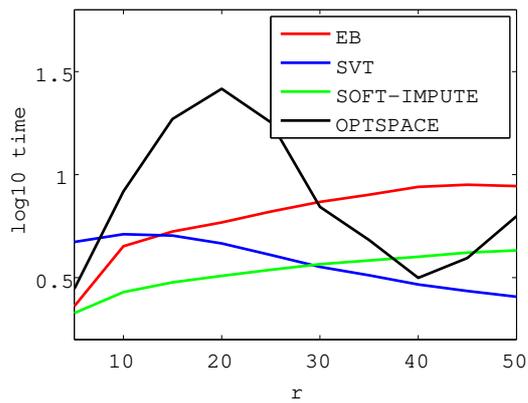}
 \end{center}
 \end{minipage}
%	\caption{Plot of (a) error2 and (b) computation time (in seconds) with respect to $r$, where $p=1000$, $q=100$, $\sigma^2=1$, and $| \Omega |/(pq)=0.5$. red: the proposed method, blue: SVT, green: SOFT-IMPUTE.}
	\caption{Plot of (a) error2 and (b) computation time (in seconds) as a function of $r$, where $p=1000$, $q=100$, $\sigma^2=1$, and $| \Omega |/(pq)=0.5$.}
	\label{rank_plot}
\end{figure}

Figure \ref{rank_plot} plots error2 and the computation time as a function of $r \in [5, 50]$, 
where $p=1000$, $q=100$, $\sigma^2=1$, and $| \Omega |/(pq)=0.5$.
Here, error2 increases with $r$ for all algorithms and EB has the best accuracy for all values of $r$.
On the other hand, the behavior of the computation time varies among the four algorithms.
Whereas the computation time of EB and SOFT-IMPUTE increases with $r$,
that of SVT decreases with $r$ and that of OPTSPACE is not monotone.
As a whole, SVT and SOFT-IMPUTE has a little better efficiency than EB and OPTSPACE.

\begin{figure}
 \begin{minipage}{0.5\hsize}
 (a)
 \begin{center}
	\includegraphics[width=7.5cm]{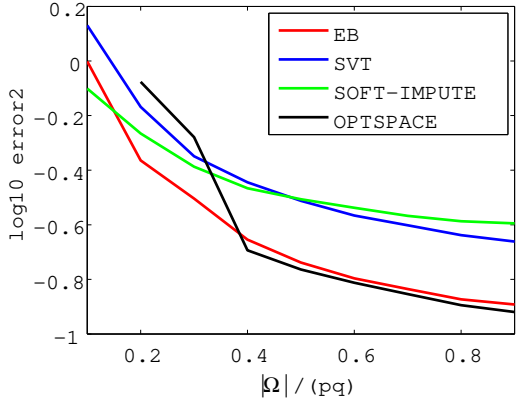}
 \end{center}
 \end{minipage}
 \begin{minipage}{0.5\hsize}
 (b)
 \begin{center}
	\includegraphics[width=7.5cm]{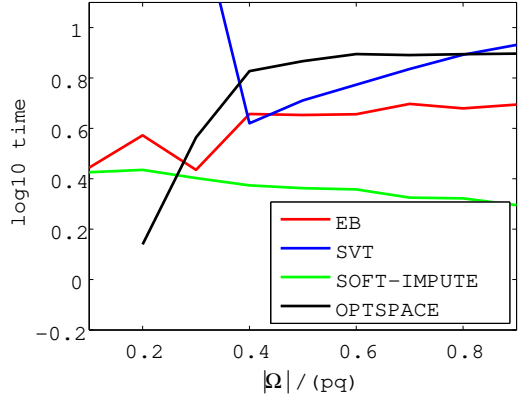}
 \end{center}
 \end{minipage}
%	\caption{Plot of (a) error2 and (b) computation time (in seconds) with respect to $| \Omega |/(pq)$, where $p=1000$, $q=100$, $r=10$, and $\sigma^2=1$. red: the proposed method, blue: SVT, green: SOFT-IMPUTE.}
	\caption{Plot of (a) error2 and (b) computation time (in seconds) as a function of $| \Omega |/(pq)$, where $p=1000$, $q=100$, $r=10$, and $\sigma^2=1$.}
	\label{obs_plot}
\end{figure}

Figure \ref{obs_plot} plots error2 and the computation time\footnote{In SVT, we set $\epsilon=10^{-3}$, since the default setting of $\epsilon=10^{-4}$ often caused divergence. When $| \Omega |/(pq) \leq 0.3$, SVT did not converge in $k_{\max}=1000$ iterations and so the computation time is extremely large. In OPTSPACE, we omit the results for $| \Omega |/(pq) = 0.1$, since the step of guessing the rank sometimes did not finish.}
as a function of $| \Omega |/(pq) \in [0.1, 0.9]$, where $p=1000$, $q=100$, $r=10$, and $\sigma^2=1$.
Here, error2 decreases with $| \Omega |/(pq)$ for all algorithms.
EB has the best accuracy for all values of $| \Omega |/(pq)$ and OPTSPACE also attains the best accuracy when $| \Omega |/(pq) \geq 0.4$.
The computation time of SOFT-IMPUTE is almost constant with $| \Omega |/(pq)$,
whereas those of EB and OPTSPACE increase with $| \Omega |/(pq)$ and almost converges at $| \Omega |/(pq) = 0.4$.
As a whole, SOFT-IMPUTE has the best efficiency and EB is the second best.

\begin{figure}
 \begin{minipage}{0.5\hsize}
 (a)
 \begin{center}
	\includegraphics[width=7.5cm]{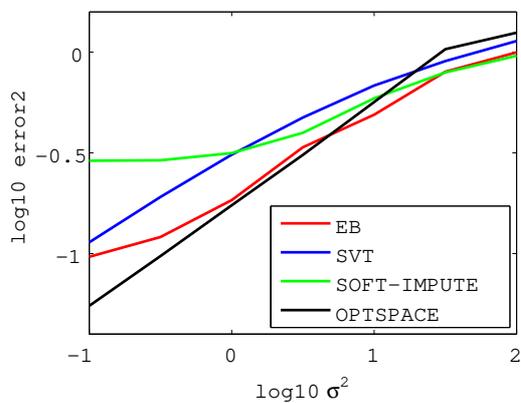}
 \end{center}
 \end{minipage}
 \begin{minipage}{0.5\hsize}
 (b)
 \begin{center}
	\includegraphics[width=7.5cm]{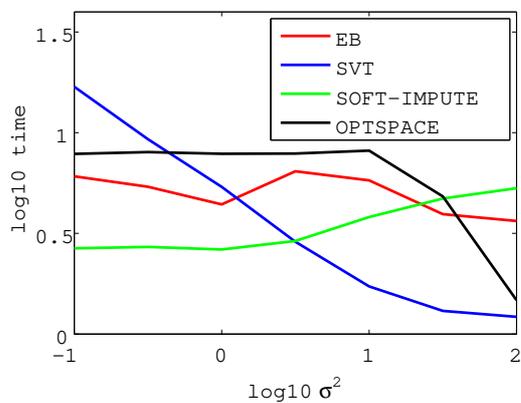}
 \end{center}
 \end{minipage}
%	\caption{Plot of (a) error2 and (b) computation time (in seconds) with respect to $\sigma^2$, where $p=1000$, $q=100$, $r=10$, and $| \Omega |/(pq)=0.5$. red: the proposed method, blue: SVT, green: SOFT-IMPUTE.}
	\caption{Plot of (a) error2 and (b) computation time (in seconds) as a function of $\sigma^2$, where $p=1000$, $q=100$, $r=10$, and $| \Omega |/(pq)=0.5$.}
	\label{noise_plot}
\end{figure}

Figure \ref{noise_plot} plots error2 and the computation time as a function of $\sigma^2 \in [10^{-1}, 10^2]$, 
where $p=1000$, $q=100$, $r=10$, and $| \Omega |/(pq)=0.5$.
For all algorithms, error2 increases with $\sigma^2$.
Whereas OPTSPACE has the best accuracy when $\sigma^2 \leq 1$, 
error2 is almost the same for all algorithms when $\sigma^2 \geq 10$. 
The computation time of EB and SOFT-IMPUTE is almost constant with $\sigma^2$,
whereas that of SVT decreases with $\sigma^2$ and that of OPTSPACE decreases rapidly when $\sigma^2 \geq 10$.

Finally, we investigate the effect of $\sigma_0^2$ on the performance of EB.
Figure \ref{init_sigma2_plot} plots error2 and the computation time of EB as a function of $\sigma_0^2 \in [10^{-2}, 10^2]$,
where $p=1000$, $q=100$, $r=10$, $\sigma^2=1$, and $| \Omega |/(pq)=0.5$.
From Figure \ref{init_sigma2_plot}, we see that error2 is almost constant with $\sigma_0^2$ except for the case $\sigma_0^2=100$.
Therefore, the accuracy of EB does not depend on $\sigma_0^2$ provided that $\sigma_0^2$ is not significantly larger than $\sigma^2$.
On the other hand, the computation time increases as $\sigma_0^2$ becomes farther from $\sigma^2$.
Interestingly, the selection $\sigma_0^2=\sqrt{10}$ provides better efficiency than the correct selection of $\sigma_0^2=\sigma^2=1$.

In summary, EB achieves a good trade-off between accuracy and efficiency 
and has comparable performance to SVT, SOFT-IMPUTE and OPTSPACE.
In particular, EB works well when the difference between the number of rows and columns is large. 
%Moreover, from the viewpoint of implementation, the proposed method does not require heuristic parameter tuning, which is inevitable in other methods.

\section{Application to Real Data}
In this section, we apply the EB algorithm to the data set from the Jester online joke recommender system \citep{Goldberg}.
This data set comprises over 4.1 million continuous ratings (-10.00 to +10.00) of $q=100$ jokes from $p=24983$ users, 
which were collected from April 1999 to May 2003 online. 
Similarly to the previous section, we compare EB with SVT, SOFT-IMPUTE, and OPTSPACE. 
In EB, we set $\varepsilon_1=10^{-3}$, $\varepsilon_2=10^{-4}$, and $\sigma_0^2=10$.
In SVT, SOFT-IMPUTE, and OPTSPACE, we adopt the default settings of the algorithm parameters.

In the original data set, $| \Omega_0 |=1810455$ among $pq=2498300$ entries were observed.
We randomly sampled $| \Omega |=5 \times 10^5$ entries from the observed entries and applied matrix completion algorithms to the sampled entries.
We evaluated the accuracy of each algorithm by the normalized error for the rest of the $| \Omega_0 |-| \Omega |=1310455$ entries:
\begin{equation*}
	{\rm error} := \frac{(\sum_{(i,j) \in \Omega_0 \setminus \Omega} (\hat{M}_{ij}-M_{ij})^2)^{1/2}}{(\sum_{(i,j) \in \Omega_0 \setminus \Omega} M_{ij}^2)^{1/2}}.
\end{equation*}

Table \ref{tab_jester} shows the results.
We only present results for EB, SOFT-IMPUTE, and OPTSPACE since SVT diverged\footnote{We also tried $E_{ij}=\hat{\sigma}$ instead of $E_{ij}=\sigma_0$, where $\hat{\sigma}^2=18.21$ is the estimate of $\sigma^2$ by EB, but it diverged again. Changing $\epsilon$ from $10^{-4}$ to $10^{-3}$ was not useful either.}.
Whereas the accuracies of the three algorithms are almost identical, EB takes the least computation time among the three algorithms.
%Although SVT is much faster than the others, its accuracy is extremely low.
This result shows the practical utility of EB on real data. % especially when $p$ is much larger than $q$.

\begin{table}[htbp]
	\centering
	\begin{tabular}{|c|c|c|}
	\hline
	 & error & time \\ \hline
	EB & 0.85 & 52.21 \\ \hline
%	SVT & 3.36 & 0.92 \\ \hline
	SOFT-IMPUTE & 0.82 & 100.11 \\ \hline
	OPTSPACE & 0.83 & 68.71 \\ \hline
	\end{tabular}
	\caption{Performance of matrix completion algorithms on the Jester data set}
	\label{tab_jester}
\end{table}

\section{Conclusion}
In this study, we proposed an empirical Bayes (EB) algorithm for matrix completion.
The EB algorithm is motivated from the Efron--Morris estimator for a normal mean matrix.
It is free from heuristic parameter tuning other than tolerance.
Numerical results demonstrated that the EB algorithm achieves a good trade-off between accuracy and efficiency compared to existing algorithms 
and that it works particularly well when the difference between the number of rows and columns is large.
In addition, application to real data showed the practical utility of the EB algorithm.

\section*{Acknowledgements}
We thank Edward I. George for helpful comments.
%This work was supported by JSPS KAKENHI Grant Numbers 14J09148.

\appendix
\section{Posterior of Normal Mean from Missing Observation}
Suppose that $m \sim {\rm N}_q (0,\Sigma)$ and $X \mid m \sim {\rm N}_q (m,\sigma^2 I_q)$.
Here, we derive the posterior distribution of $m$ given a subvector of $X$.

First, consider the case where we observe the first $k$ entries of $X$, which we denote by $Z=(X_1,\cdots,X_k)$.
Let $C \in \mathbb{R}^{k \times k}$ be defined by
\begin{equation*}
	C = (\sigma^2 I_k + \Sigma[\gamma_k,\gamma_k])^{-1}, \quad \gamma_k = \{ 1,\cdots,k \}.
\end{equation*}
Then, the posterior distribution of $m$ given $Z=z$ is obtained as
\begin{align*}
	p(m \mid Z=z) &\propto \pi(m) p(Z=z \mid m) \\
	&\propto \exp \left( -\frac{1}{2} m^{\top} \Sigma^{-1} m -\frac{\sum_{i=1}^k (z_i-m_i)^2}{2 \sigma^2} \right) \\
	&\propto \exp \left( -\frac{1}{2} (m-\hat{m})^{\top} R^{-1} (m-\hat{m}) \right),
\end{align*}
where
\begin{equation*}
	R = \left( \Sigma^{-1}  + \sigma^{-2} \begin{pmatrix} I_k & O \\ O & O \end{pmatrix} \right)^{-1} = \Sigma-\Sigma \begin{pmatrix} C & O \\ O & O \end{pmatrix} \Sigma,
\end{equation*}
\begin{equation*}
	\hat{m} = \sigma^{-2} R \begin{pmatrix} z \\ 0 \end{pmatrix}.
\end{equation*}
Here, we used the Sherman-Morrison-Woodbury formula \citep{Golub}
\begin{equation*}
	(\Sigma^{-1}+U^{\top}U)^{-1} = \Sigma-\Sigma U^{\top} (I+ U \Sigma U^{\top})^{-1} U \Sigma.
\end{equation*}
Therefore, the posterior distribution of $m$ given $Z=z$ is the normal distribution ${\rm N}(\hat{m}, R)$.

The above result is straightforwardly extended to the general case 
where we observe $k$ entries of $X$ with indices $\omega=\{ \omega_1,\cdots, \omega_k \} \subset \{ 1,\cdots,q \}$. 
We denote the observed subvector of $X$ by $Z=(X_{\omega_1},\cdots,X_{\omega_k})$.
Let ${P} \in \mathbb{R}^{q \times q}$ and $b \in \mathbb{R}^q$ be defined by
\begin{equation*}
	{P} [\omega,\omega] = (\sigma^2 I_k + \Sigma[\omega,\omega])^{-1}, \quad {P}_{j_1 j_2} = 0 \ (j_1 \not\in \omega \ {\rm or} \ j_2 \not\in \omega),
\end{equation*}
\begin{equation*}
	b_j = \begin{cases} Z_l & (j=\omega_l) \\ 0 & (j \not\in \omega) \end{cases}.
\end{equation*}
Then, the posterior distribution of $m$ given $Z=z$ is obtained as
\begin{align*}
	p(m \mid Z=z) &\propto \exp \left( -\frac{1}{2} (m-\hat{m})^{\top} R^{-1} (m-\hat{m}) \right),
\end{align*}
where
\begin{equation*}
	R = \Sigma - \Sigma {P} \Sigma,
\end{equation*}
\begin{equation*}
	\hat{m} = \sigma^{-2} R b.
\end{equation*}
Therefore, the posterior distribution of $m$ given $Z=z$ is the normal distribution ${\rm N}(\hat{m}, R)$.

\section{Derivation of the EB Algorithm}
In general, the EM algorithm is used to iteratively estimate the parameters of statistical models with latent variables $p(X,Z \mid \theta)$ \citep{EMalgo}.  
Here, $X$ denotes the observed variables and $Z$ denotes the latent variables.
Each iteration in the EM algorithm is described as
\begin{equation*}
	\theta_{t+1} = {\rm arg} \max_{\theta} Q(\theta, \theta_t), 
\end{equation*}
where
\begin{equation*}
	Q(\theta, \theta_t) = {\rm E}_{Z \mid X,\theta_t} [\log p(X,Z \mid \theta)].
\end{equation*}

In the present model \eqref{gen_model} and \eqref{obs_model}, $\theta=(\Sigma,\sigma^2)$ is the parameter, 
$Y_{\Omega}$ is the observed variables, and $M$ is the latent variables.
The log-likelihood function is calculated as
\begin{align*}
	\log p(Y_{\Omega},M \mid \theta) &= \log p(M \mid \Sigma) + \log p(Y_{\Omega} \mid M, \sigma^2) \\
	&= -\frac{p}{2} \log \det \Sigma -\frac{1}{2} {\rm tr} \Sigma^{-1} M M^{\top} -\frac{|\Omega|}{2} \log \sigma^2 -\frac{1}{2 \sigma^2} \sum_{(i,j) \in \Omega} (Y_{ij}-M_{ij})^2.
\end{align*}
From the covariance structure in \eqref{gen_model} and \eqref{obs_model}, 
row vectors $m_1,\cdots,m_p$ of $M=(m_1,\cdots,m_p)^{\top}$ are independent given $Y_{\Omega}$.
Using the results in Appendix A, the posterior distribution of each $m_i$ given $Y_{\Omega}$ is obtained as
\begin{align*}
	m_i \mid Y_{\Omega}, \theta \sim {\rm N} (\hat{m}_i, R_i),% \quad m_i \indepe m_j \ (i \neq j),
\end{align*}
where
\begin{equation*}
	R_i = \Sigma-\Sigma {P}_i \Sigma,
\end{equation*}
\begin{equation*}
	\hat{m}_i = \sigma^{-2} R_i b_i.
\end{equation*}
Here, ${P}_i \in \mathbb{R}^{q \times q}$ and $b_i \in \mathbb{R}^q$ are defined by
\begin{equation*}
	{P}_i [\Omega_i,\Omega_i] = (\sigma^2 I_{|\Omega_i|} + \Sigma [\Omega_i,\Omega_i])^{-1}, \quad ({P}_i)_{j_1 j_2} = 0 \quad (j_1 \not\in \Omega_i \ {\rm or} \ j_2 \not\in \Omega_i),
\end{equation*}
\begin{equation*}
	(b_i)_j = \begin{cases} Y_{i j} & (j \in \Omega_i) \\ 0 & (j \not\in \Omega_i) \end{cases}.
\end{equation*}

Therefore,
\begin{align*}
	Q(\widetilde{\theta}, \theta) &= {\rm E}_{M \mid Y_{\Omega}, \theta} [\log p(Y_{\Omega}, M \mid \widetilde{\theta})] \\
	&= Q_1 (\widetilde{\Sigma}, \Sigma) + Q_2 (\widetilde{\sigma}^2, \sigma^2),
\end{align*}
where
\begin{equation*}
	Q_1(\widetilde{\Sigma}, \Sigma) = -\frac{p}{2} \log \det \widetilde{\Sigma} -\frac{1}{2} \sum_{i=1}^p {\rm tr} \left( \widetilde{\Sigma}^{-1} \hat{m}_i \hat{m}_i^{\top} + R_i \right),
\end{equation*}
\begin{equation*}
	Q_2(\widetilde{\sigma}^2, \sigma^2) = -\frac{|\Omega|}{2} \log \widetilde{\sigma}^2 -\frac{1}{2 \widetilde{\sigma}^2} \sum_{(i,j) \in \Omega} ((Y_{ij}-(\hat{m}_i)_j)^2 + (R_i)_{jj}).
\end{equation*}
By maximizing $Q_1(\widetilde{\Sigma}, \Sigma)$ and $Q_2(\widetilde{\sigma}^2, \sigma^2)$ with respect to $\widetilde{\Sigma}$ and $\widetilde{\sigma}^2$ respectively, we obtain
\begin{equation*}
	\widetilde{\Sigma} = \frac{1}{p} \sum_{i=1}^p \left( \hat{m}_i \hat{m}_i^{\top} + R_i \right),
\end{equation*}
\begin{equation*}
	\widetilde{\sigma}^2 = \frac{1}{|\Omega|} \sum_{(i,j) \in \Omega} \left( (Y_{ij}-(\hat{m}_i)_j)^2 + (R_i)_{jj} \right).
\end{equation*}
Thus, Algorithm \ref{EBalgo} is obtained.


\begin{thebibliography}{99}
\expandafter\ifx\csname natexlab\endcsname\relax\def\natexlab#1{#1}\fi

\bibitem[{Cai, Cand\`{e}s and Shen(2010)}]{Cai}
\textsc{Cai, J. F.}, \textsc{Cand$\grave{{\rm e}}$s, E. J.} \& \textsc{Shen, Z.} (2010).
\newblock{A singular value thresholding algorithm for matrix completion}.
\newblock \textit{SIAM Journal on Optimization} \textbf{20}, 1956--1982.

\bibitem[{Cand\`{e}s and Recht(2008)}]{Candes}
\textsc{Cand$\grave{{\rm e}}$s, E. J.} \& \textsc{Recht, B.} (2008).
\newblock{Exact matrix completion via convex optimization}.
\newblock \textit{Foundations of Computational Mathematics} \textbf{9}, 717--772.

\bibitem[{Dawid(1981)}]{Dawid}
\textsc{Dawid, A. P.} (1981).
\newblock{Some matrix-variate distribution theory: notational considerations and a Bayesian application}.
\newblock \textit{Biometrika} \textbf{68}, 265--274.

\bibitem[{Dempster, Laird and Rubin(1972)}]{EMalgo}
\textsc{Dempster, A. P.}, \textsc{Laird, N. M.} \& \textsc{Rubin, D. B.} (1972).
\newblock{Maximum likelihood from incomplete data via the EM algorithm}.
\newblock \textit{Journal of the Royal Statistical Society B} \textbf{39}, 1--38.

\bibitem[{Efron and Morris(1972)}]{Efron72}
\textsc{Efron, B.} \& \textsc{Morris, C.} (1972).
\newblock{Empirical Bayes on vector observations: an extension of Stein's method}.
\newblock \textit{Biometrika} \textbf{59}, 335--347.

\bibitem[{Goldberg et al.(2001)}]{Goldberg}
\textsc{Goldberg, K.}, \textsc{Roeder, T.}, \textsc{Gupta, D.} \& \textsc{Perkins, C.} (2001).
\newblock{Eigentaste: A constant time collaborative filtering algorithm}.
\newblock \textit{Information Retrieval} \textbf{4}, 133--151.

\bibitem[{Golub and van Loan(1996)}]{Golub}
\textsc{Golub, G. H.} \& \textsc{van Loan, C. F.} (1996).
\newblock \textit{Matrix Computations}.
\newblock Baltimore, MD: Johns Hopkins.

\bibitem[{Keshavan, Montanari, and Oh(2010)}]{Keshavan}
\textsc{Keshavan, R. H.}, \textsc{Montanari, A.} \& \textsc{Oh, S.} (2010).
\newblock{Matrix completion from noisy entires}.
\newblock \textit{Journal of Machine Learning Research} \textbf{11}, 2057--2078.

\bibitem[{Matsuda and Komaki(2015)}]{Matsuda}
\textsc{Matsuda, T.} \& \textsc{Komaki, F.} (2015).
\newblock{Singular value shrinkage priors for Bayesian prediction}.
\newblock \textit{Biometrika} \textbf{102}, 843--854.

\bibitem[{Mazumder, Hastie and Tibshirani(2010)}]{Mazumder}
\textsc{Mazumder, R.}, \textsc{Hastie, T.} \& \textsc{Tibshirani, R.} (2010).
\newblock{Spectral regularization algorithms for learning large incomplete matrices}.
\newblock \textit{Journal of Machine Learning Research} \textbf{11}, 2287--2322.

\bibitem[{Recht(2011)}]{Recht}
\textsc{Recht, B.} (2011).
\newblock{A simpler approach to matrix completion}.
\newblock \textit{Journal of Machine Learning Research} \textbf{12}, 3413--3430.

\bibitem[{Srebro(2005)}]{Srebro}
\textsc{Srebro, N.}, \textsc{Rennie, J.} \& \textsc{Jaakkola, T.} (2005).
\newblock{Maximum-margin matrix factorization}.
\newblock In \textit{Advances in Neural Information Processing Systems} \textbf{17}, 1329--1336.

\bibitem[{Stein(1974)}]{Stein74}
\textsc{Stein, C.} (1974).
\newblock{Estimation of the mean of a multivariate normal distribution}.
\newblock \textit{Proceedings of Prague Symposium on Asymptotic Statistics} \textbf{2}, 345--381.
\end{thebibliography}
\end{document}